# A Comparative Study of GAN-Generated Handwriting Images and MNIST Images using t-SNE Visualization

Okan Düzyel[1]

[1]Department of Electrical and Electronics Engineering
Izmir Institute of Technology

okanduzyel@iyte.edu.tr

**Abstract**

*The quality of GAN-generated images on the MNIST dataset was explored in this paper by comparing them to the original images using t-distributed stochastic neighbor embedding (t-SNE) visualization. A GAN was trained with the dataset to generate images and the result of generating all synthetic images, the corresponding labels were saved. The dimensionality of the generated images and the original MNIST dataset was reduced using t-SNE and the resulting embeddings were plotted. The rate of the GAN-generated images was examined by comparing the t-SNE plots of the generated images and the original MNIST images. It was found that the GAN-generated images were similar to the original images but had some differences in the distribution of the features. It is believed that this study provides a useful evaluation method for assessing the quality of GAN-generated images and can help to improve their generation in the future.*

## 1. Introduction

The MNIST (Modified National Institute of Standards and Technology) is a widely preferred dataset for classification algorithms, noise reduction, generative models, etc. Since it is its introduction by LeCun et al. in 1998 [1]. The entire dataset contains 70,000 trains and 10,000 test data. Each image is grayscale and has a resolution of 28x28 pixels. The dataset has been used and is still used in many applications where artificial intelligence algorithms are used. These algorithms are convolutional neural networks (CNN), dense neural networks (DNN), support vector machines (SVM), etc.

One of the main challenges in machine learning is generating new synthetic data that is similar to the original dataset. This is particularly relevant in scenarios where collecting new data is time-consuming or expensive. Generative Adversarial Networks (GANs) have emerged as a powerful tool for generating synthetic data [2]. GANs consist of two distinct parts, which are named a generator and a discriminator parts, which have trained each other (compete with each other). While the generator tries to generate new data, the discriminator tries to classify the real or fake accuracy of this data. During the entire training, these two networks compete with each other, and the models are trained by updating the weights of the two models in each trial. GAN can be used not only to produce images but also to produce audio signals, tabular data, text, etc.

The aim of this paper is to investigate the use of GANs for generating synthetic MNIST data and to compare the results of GAN-generated data with the original MNIST dataset using t-SNE. Specifically, aiming to explore the effectiveness of GANs for generating new synthetic data that is similar to the original dataset. t-SNE provides information about the similarity of the existing data stochastically which data belong to any classes by showing the use of the distances between the data in the two-dimensional or three-dimensional plane. Sometimes it is sufficient to examine the results in two dimensions, but in some cases, the representation in two dimensions is insufficient because of dimensional representation inability, and the results may be necessary to investigate under the three-dimensional results.

This paper is organized as follows. Section 2 provides detailed background on MNIST and GANs. Section 3 presents related work in the field of MNIST and GANs. Section 4 describes the methodology used in this study, including the architecture of the GAN, the training process, and the evaluation of the generated data. Section 5 presents the results of the experiments, including a comparison of the synthetic data generated by the GAN with the original MNIST dataset using t-SNE. Finally, Section 6 presents the conclusions of the study and discusses future work.

## 2. Background of MNIST dataset and GAN

The MNIST dataset is widely used for developing and evaluating algorithms for image recognition, classification, and other computer vision tasks. It has been used extensively in the development of deep learning models such as convolutional neural networks (CNNs) for image recognition and is often used as a standard dataset in online courses and tutorials for machine learning and computer vision.

Tremendous successes have been achieved in CNN-based studies on the MNIST dataset. According to the study, with top models achieved over 99% accuracy on the test set [3]. However, as with any machine learning dataset, the performance of the models is limited by the size and quality of the dataset. In such cases, data needs to be replicated and there are two methods of data augmentation. Classical data augmentation methods [4] could be for images, rotation, zoom, flipping adding noise, etc. These methods only change the existing data, they cannot synthesize new data. GAN can synthesize new data, and with this newly synthesized data, the performance of the machine learning model can be increased. GANs, or Generative Adversarial Networks, is a type of deep learning model that has gained significant popularity for their ability to generate synthetic data that closely resembles real data. The random generator module and generator network try to generate new synthetic data and the discriminator network decides if it is fake or real. The two networks compete with each other in an adversarial manner, with the generator attempting to trick the discriminator, and the task of the discriminator that needs to make the correct decision about generated new data, fake or real.

The GAN architecture can be applied to many different areas, including text generation, tabular data generation, and more. GANs have been shown to be effective at generating new synthetic data that is like the original dataset and can be used to augment existing datasets or generate entirely new datasets. In the context of the MNIST dataset, GANs have been used to generate synthetic handwritten digits that are visually similar to the real handwritten digits in the MNIST dataset.

t-SNE (t-Distributed Stochastic Neighbor Embedding) is a technique for visualizing high-dimensional data in a lower-dimensional space [5]. t-SNE is particularly useful for visualizing clusters and patterns in the data and can be used to compare the similarities and differences between two datasets. In the context of the MNIST dataset, t-SNE can be used to visualize the similarities and differences between the original MNIST dataset, and the synthetic data generated by the GAN.

In this paper, we aim to investigate the use of GANs for generating synthetic MNIST data, and to compare the results of GAN-generated data with the original MNIST dataset using t-SNE. The results of this study will provide insights into the effectiveness of GANs for generating new synthetic data that is similar to the original dataset and will provide a valuable tool for evaluating the quality of the generated data.

## 3. Related Works

MNIST has been used extensively as a benchmark dataset for evaluating machine learning models. Over the years, a wide range of models has been evaluated on the MNIST dataset, including support vector machines (SVMs) [6], decision trees [7], and convolutional neural networks (CNNs) [3]. Among these models, CNNs have been shown to achieve state-of-the-art performance on the MNIST dataset, with top models achieving over 98% accuracy on the test set [8].

In recent years, generative adversarial networks (GANs) have emerged as powerful tools for generating synthetic data. Shortly before the GAN, the Variational Autoencoder (VAE) [9] was announced. Unlike the traditional Autoencoder, VAE transfers the data from the latent space (bottleneck) part to the decoder according to the mean and variance and resulting in this process, new data could be synthesized. In the context of the MNIST dataset, GANs have been used to generate synthetic handwritten digits that are visually similar to the real handwritten digits in the MNIST dataset. For example, Radford et al. [10] used a deep convolutional GAN to generate synthetic MNIST digits and showed that the generated digits were visually like the real MNIST digits. Similarly, Salimans et al. [11] used a GAN to generate synthetic MNIST digits and showed that the generated digits could be used to augment the original MNIST dataset and improve the performance of a CNN on the task of digit recognition.

t-SNE has also been used extensively in the field of machine learning for visualizing high-dimensional data. In the context of the MNIST dataset, t-SNE has been used to visualize the similarities and differences between different models and representations of the data. For example, van der Maaten and Hinton [6] used t-SNE to visualize the similarities and differences between different models of the MNIST dataset and showed that the t-SNE visualization could be used to identify patterns and clusters in the data.

In this paper, we aim to build upon the work of previous researchers and investigate the use of GANs for generating synthetic MNIST data and compare the results of GAN-generated data with the original MNIST dataset using t-SNE. Our study will provide further insights into the effectiveness of GANs for generating new synthetic data that is similar to the original dataset and will provide a valuable tool for evaluating the quality of the generated data.

# 4. Methodology

## 4.1. GAN

Generative Adversarial Networks (GANs) is an unsupervised model, and its purpose is to synthesize new data with the data it learns from the dataset. It does this by competing for two different neural networks in it with each other. As a result of this competition, the discriminator can no longer distinguish between synthetic and real data, and the generator wins the game. In other words, this is the point where the loss functions of the generator and the discriminator intersect, and this point is called the Nash equilibrium point. If both loss functions are defined between (e.g.) 0-1 and during the training period, as the loss rate of one function increases, that of the other function decreases. Therefore, Nash equilibrium usually occurs at the midpoint, at a value of 0.5.

The generator network takes a random noise vector z as input and generates a new sample x' that is similar to the training data. The discriminator network takes either a real sample x from the training dataset or a fake sample x' generated by the generator network as input and outputs a scalar value $D(x)$ which represents the probability that the input sample is real. During training, the generator tries to generate synthetic samples that the discriminator is unable to differentiate from real samples, while the discriminator tries to correctly identify whether the samples are real or fake. The training process is formalized as a game, where the generator tries to decrease the objective function $L_G$ while the discriminator tries to increase it. The objective of all GAN's functions is defined as:

$$L_D = Error(D(x),1) + Error(D(G(z)),0) \quad (4.1.1)$$

$$L_G = Error(D(G(z)),1) \quad (4.1.2)$$

$$H(p,q) = \mathbb{E}_{x \sim p(x)}[-\log q(x)] \quad (4.1.3)$$

$$\min_G \max_D V(D,G) = \mathbb{E}_{x \sim p_{data}(x)}[\log D(x)] + \mathbb{E}_{z \sim p_z(z)}[\log(1 - D(g(z)))] \quad (4.1.4)$$

Where $p_{data}(x)$ is the distribution of real data, $p_z(z)$ is the distribution of random noise vectors, $G(z)$ is the generator network, and $D(x)$ is the discriminator network. The first term in the objective function represents the expected value of the logarithm of the discriminator output for real data, while the second term represents the expected value of the logarithm of the discriminator output for fake data generated by the generator. The generator tries to minimize this objective function, while the discriminator tries to maximize it.

## 4.2. t-SNE

t-SNE is a mathematical approach for representing a multidimensional data in two- or three-dimensional space. This method is applied in data science and engineering applications to determine the similarities or differences of existing data and to determine which data belongs to which class or clusters if the data don't have labels via representing with creating space between representation points in lower dimensional space.

Let's assume that we have a dataset with n data points and a similarity matrix $S$, where $Sij$ represents the similarity between data point $i$ and $j$.

First, we start by computing a similarity matrix $P$ in the multi-dimensional space, where each element $pi|j$ represents the conditional probability of x data point $i$ as a neighbor of data point j, given the Gaussian kernel centered at $j$. The formula for $pi|j$ is:k

$$p_{j|i} = \frac{e^{\frac{-\|x_i - x_j\|^2}{2\sigma_i^2}}}{\sum_{k \neq i}^{N} e^{\frac{-\|x_i - x_k\|^2}{2\sigma_i^2}}} \quad (4.2.1)$$

Where $xi$ and $xj$ represent data points in the high-dimensional space, $\sigma i$ is the variance of the Gaussian that is centered on data point $i$, and the denominator is a normalization constant that ensures that $\Sigma j \, pij = 1$.

Second, we compute a similarity matrix $Q$ in the low-dimensional space, where each element $qij$ represents the joint probability of finding data point i and j as neighbors, assuming a student t-distribution with one degree of freedom. The formula for $qij$ is:

$$q_{j|i} = \frac{(1 + \|y_i - y_j\|^2)^{-1}}{\sum_{k \neq i}^{N}(1 + \|y_i - y_k\|^2)^{-1}} \quad (4.2.2)$$

where $yi$ and $yj$ are the corresponding points in the low-dimensional space.

Third, we now need to compute the KL divergence between $P$ and $Q$, which measures the difference between the two distributions. The formula for the KL divergence is:

$$D_{KL}(P \| Q) = \sum_{j=1}^{N} P(x) \log\left(\frac{P(x)}{Q(x)}\right) \quad (4.2.3)$$

Low-dimensional embedding to minimize the KL divergence needs to be updated. In order to reduce KL divergence, the gradient descent method is used for this process and the positions in the low-dimensional space are updated with the help of the formula (4.2.4).

$$y_i(t+1) = y_i(t) + \eta \sum_{j=1}^{N}(p_{ij} - q_{ij})(y_i - y_j)(1 + \|y_i - y_j\|^2)^{-1} \quad (4.2.4)$$

where $\eta$ is the learning rate, t is the iteration number, and $y_i(t)$ and $y_i(t+1)$ are the $i-th$ data point's positions in the lower-dimensional space before and after the update, respectively.

**4.3. Hyperparameter tuning for the training of the model**

The MNIST dataset's image size is 28x28 pixels and the CSV version of the dataset is represented by 1x784 pixels (The size of actual data is 785. The first column represents labels of each row) for each image. Thus, the input of the discriminator and the output of the generator has selected these dimensions. Architectures of the discriminator and generator are visualized in Fig.1. TensorFlow framework is selected to implement the GAN model, which consisted of the generator network and the discriminator. Random noise is created by a random noise generator and the generator network accepts it as input and then produces a synthetic image of a handwritten digit. The discriminator network grasps an image either real or synthetic.

The Discriminator part is a model that makes binary classification at the end of the decreasing number of neurons like standard MLP models. It resembles deep learning models thanks to its hidden layers. The generator part accepts a random noise generator as input and in this study, the size of the random noise generator input is 1x100. Discriminator, Generator has a structure that grows layer size and as a result, it reaches image size. This means that the result of the generator network will be a complete image.

GAN is trained for 2400 epochs, the optimizer is selected as the "Adam" for both models, and the learning rate is selected as 0.0002 and a batch size of 128 hyperparameters. After training, 4,800 synthetic images were generated with GAN.

To evaluate the quality of the GAN-generated data, we used t-SNE, a powerful tool for visualizing high-dimensional data. We visualized the 4,800 synthetic images and 4,800 randomly selected images from the MNIST test dataset using t-SNE and compared the resulting visualizations to identify similarities and differences between the two datasets.

In summary, our methodology involved training a GAN model to generate synthetic handwritten digits and visualizing the GAN-generated data and the MNIST dataset using t-SNE.

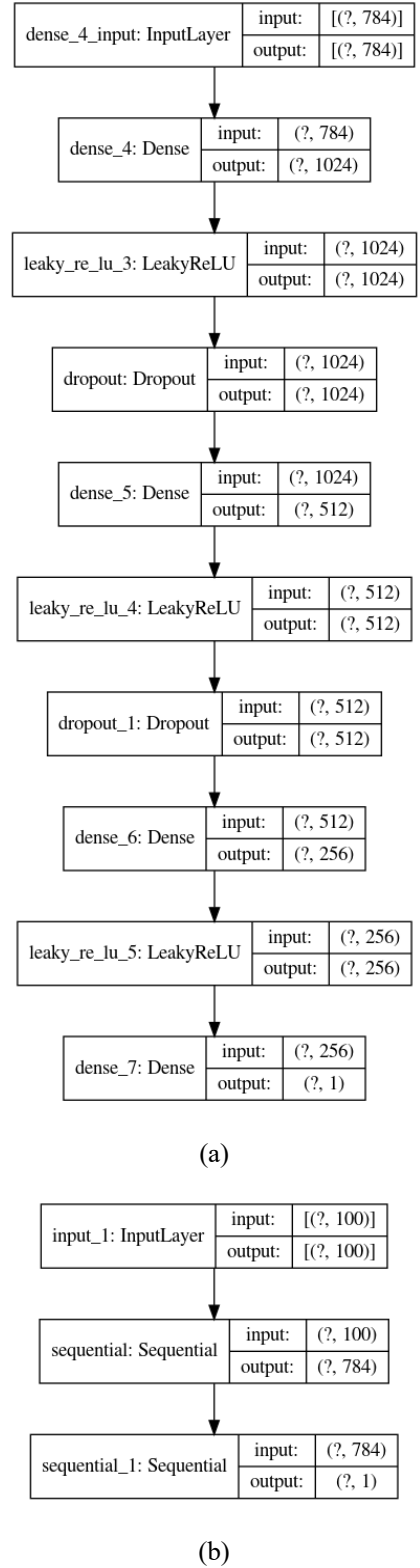

Fig.1 Architecture of discriminator (a) and generator (b) parts of GAN

## 5. Results

During the initial epoch of GAN training, the synthetic data produced by the generator is usually random noise, as the generator has not yet learned to create realistic samples that resemble the training data. However, as the training continues, the synthetic data produced by the generator gradually improves in quality, due to the generator learning from the feedback it receives from the discriminator, and the discriminator becoming more adept at distinguishing between real and synthetic data. At the end of these processes, the outputs look more similar to the original data. The training process of GAN should have made decisions according to the loss functions output of the generator and discriminator.

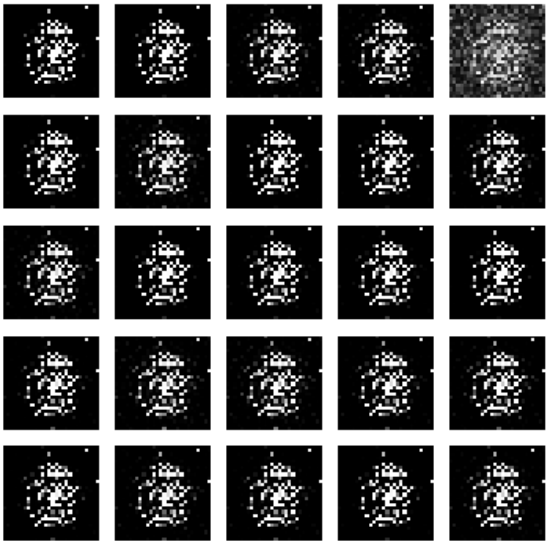

Fig.2 (a)

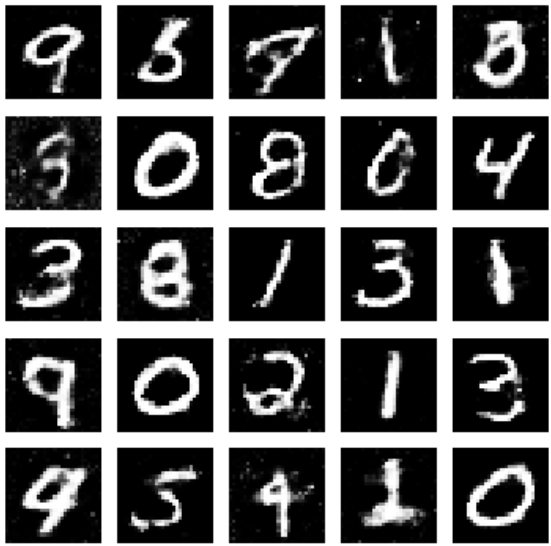

Fig.2 (b)

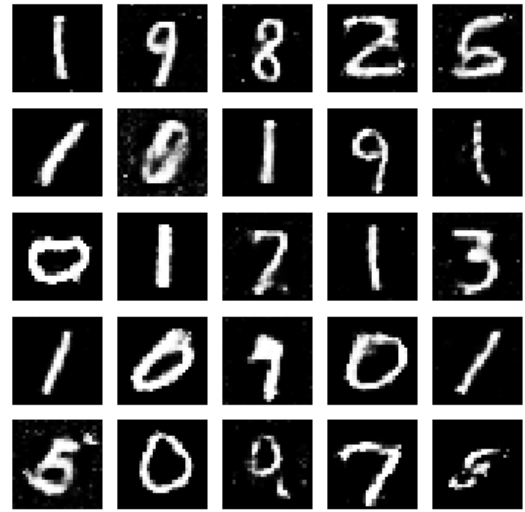

Fig.2 (c)

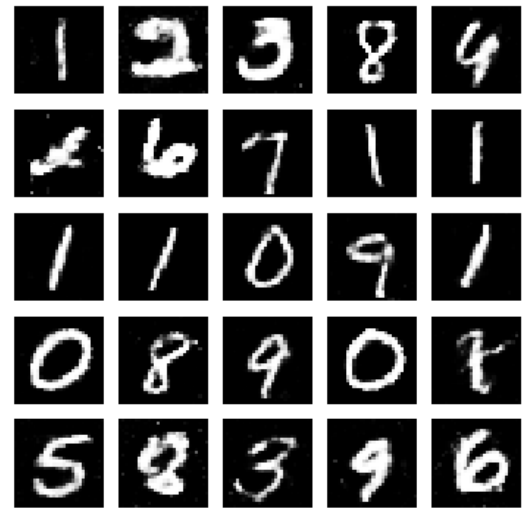

Fig.2 (d)

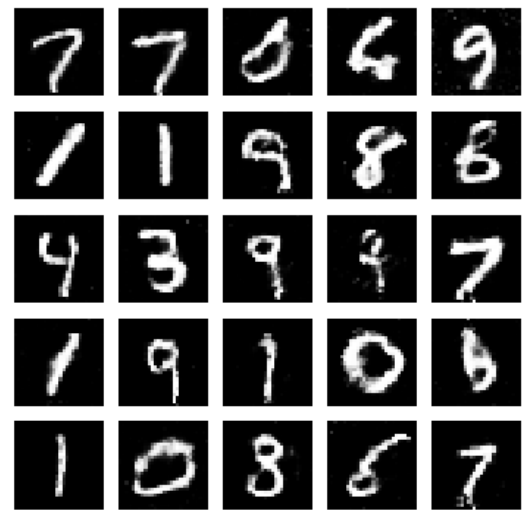

Fig.2 (e)

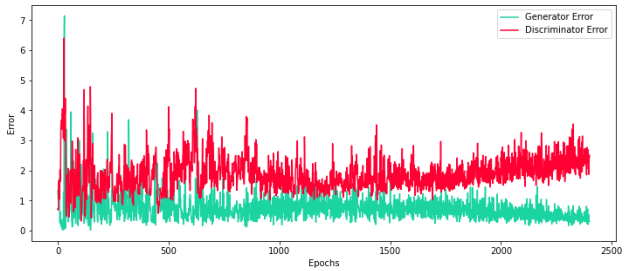

Fig.3

In a particular study, the study generated 100 synthetic data samples after every 50 epochs and selected random 25 samples from different epochs to analyze the quality of the synthetic data produced by the GAN over the course of the training. The selected epochs included the first epoch, the 300th epoch, the 600th epoch, the 1200th epoch, and the 2400th epoch. By examining the quality of the synthetic data produced at different epochs, the researchers can observe how the GAN improves its ability to produce realistic synthetic data over time. All outputs are shown in Fig.2.

If the results are examined carefully, it will be noticed that as we approach the end of the training, the synthesized data are very close to the actual data. This difference can be easily noticed between Fig.2(a) and Fig.2(e). However, these results, which we think are close to reality, should be analyzed mathematically. For this, first of all, it is necessary to look at the t-SNE results of the MNIST dataset for all classes. It is shown in Fig.4. Since the image that will be created here contains 70,000 data, it can only give us limited information visually. Therefore, it is necessary to select a specific class and examine only that class and the new data generated from this class.

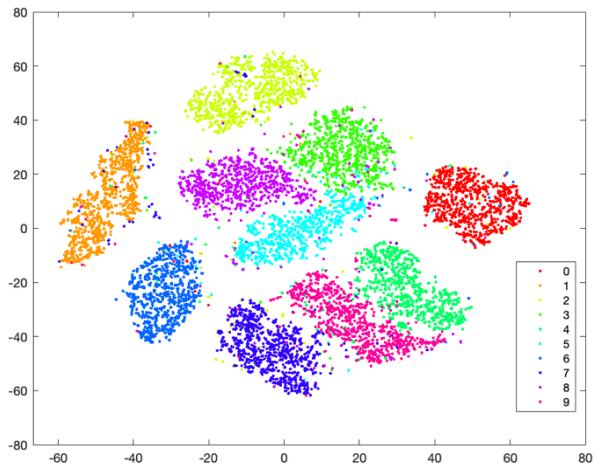

Fig.4 t-SNE result of MNIST dataset in 2D

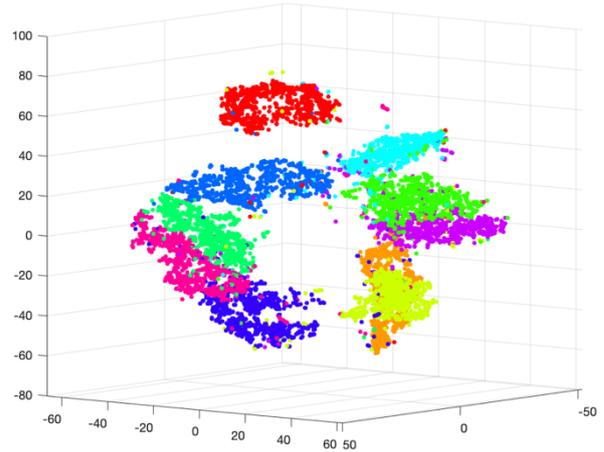

Fig.5 t-SNE result of MNIST dataset in 3D

t-SNE works according to the dimensionality reduction logic. The results can be in two dimensions or in three dimensions. Two-dimensional results sometimes do not provide enough information or may be deceptive. Because the data that seems to be intertwined in two dimensions are perhaps easily separable in the third dimension. For example, green and turquoise color classes appear to be mixed in 2D results, while in 3D results there is a gap between them. Therefore, it is necessary to examine the 3D results. 3D t-SNE results of all classes of the MNIST dataset are given in Fig.5.

Colored cloud-like views represent classes. If the figure is examined in detail, it can be observed that there are a few different colors in some classes. These data mean that they are more similar to different classes, not to the class they actually belong to. This causes the neural network model to fail in applications such as classification.

Although data was generated from each class in this study, the t-SNE results were displayed on only one class. The reason for this is that it is not possible to visually distinguish between 70,000 and synthetic data in the small size figure. Therefore, the class belonging to the number 5 was chosen randomly, and only 3000 data were used in this class in terms of visual uniformity. The t-SNE result in the last case is shown in Fig.6.

Since t-SNE stochastically reduces the data to two or three dimensions that can be visually distinguished from the N dimension, it is quite normal to observe differences in the results when adding or subtracting from the existing data. When 750 synthetic data is added to the results obtained in Fig.6, the new result is as in Fig.7. The number of synthetic data is kept low in order to create clear visuals. Even more data than class size can be produced. Fig.7 tells us is that the synthetic data (red) added to the original data (blue) are not located far from each other, but nested. This means that the generated data and the original data are very similar to each other and they can represent each other very well.

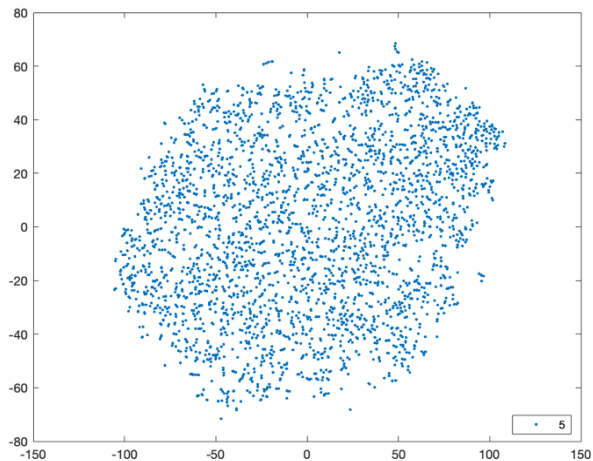
Fig.6 t-SNE result of real data from Class 5

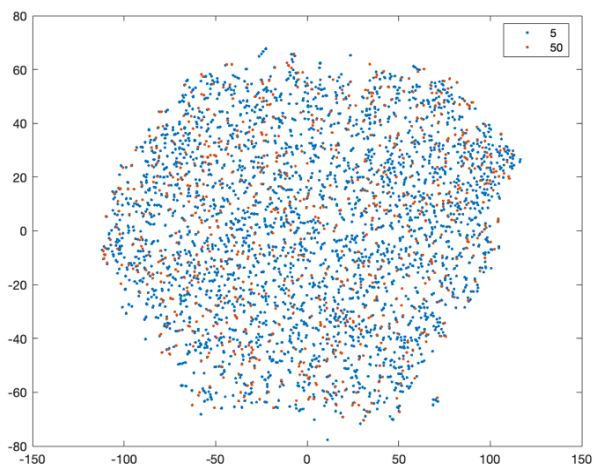
Fig.7 t-SNE result of real and synthetic data from Class 5

## 6. Conclusion

In this paper, generated synthetic data from the MNIST dataset have investigated the use of generative adversarial networks (GANs) for generating synthetic handwritten digits that are visually similar to the digits in the original dataset. Our results show that GANs can be used to generate synthetic data that is visually similar to the original MNIST dataset. We have also shown that t-SNE can be used to visualize the similarities between original and GAN-generated data.

This study has several implications for the field of machine learning. First, our results demonstrate the potential for using GANs to generate new synthetic data. Second, our study highlights the importance of using visualization tools such as t-SNE to evaluate the quality of generated data and to identify patterns and clusters in high-dimensional data.

In conclusion, our study provides valuable insights into the use of GANs for generating synthetic data from the MNIST dataset. Instead of using classical data augmentation methods, this study can serve as a foundation for further research on the use of GANs and other generative models for data augmentation and generation.